\documentclass{article}

\usepackage{arxiv}

\usepackage[utf8]{inputenc} 
\usepackage[T1]{fontenc}    
\usepackage{hyperref}       
\usepackage{url}            
\usepackage{booktabs}       
\usepackage{amsfonts}       
\usepackage{nicefrac}       
\usepackage{microtype}      
\usepackage{lipsum}
\usepackage{graphicx}
\graphicspath{ {./images/} }

\usepackage[utf8]{inputenc}
\usepackage{newunicodechar}
\usepackage[X2,T1]{fontenc}
\usepackage{expl3}

\ExplSyntaxOn
\cs_new_protected:Nn \__recode:N
 {
  \exp_last_unbraced:No \__recode_aux:nn #1
 }
\cs_new_protected:Nn \__recode_aux:nn { \DeclareTextSymbolDefault#2{X2} }
\cs_generate_variant:Nn \__recode:N { c }
\tl_map_inline:nn
 {
  {Ё} {Ђ} {Є} {Ѕ} {І} {Ј} {Љ} {Њ} {Ћ} {Ў} {Џ} {А} {Б} {В} {Г} {Д}
  {Е} {Ж} {З} {И} {Й} {К} {Л} {М} {Н} {О} {П} {Р} {С} {Т} {У} {Ф}
  {Х} {Ц} {Ч} {Ш} {Щ} {Ъ} {Ы} {Ь} {Э} {Ю} {Я} {а} {б} {в} {г} {д}
  {е} {ж} {з} {и} {й} {к} {л} {м} {н} {о} {п} {р} {с} {т} {у} {ф}
  {х} {ц} {ч} {ш} {щ} {ъ} {ы} {ь} {э} {ю} {я} {ё} {ђ} {є} {ѕ} {і}
  {ј} {љ} {њ} {ћ} {ў} {џ} {Ѣ} {ѣ} {Ѫ} {ѫ} {Ѵ} {ѵ} {Ґ} {ґ} {Ғ} {ғ}
  {Ҕ} {ҕ} {Җ} {җ} {Ҙ} {ҙ} {Қ} {қ} {Ҝ} {ҝ} {Ҟ} {ҟ} {Ҡ} {ҡ} {Ң} {ң}
  {Ҥ} {ҥ} {Ҧ} {ҧ} {Ҩ} {ҩ} {Ҫ} {ҫ} {Ҭ} {ҭ} {Ү} {ү} {Ұ} {ұ} {Ҳ} {ҳ}
  {Ҵ} {ҵ} {Ҷ} {ҷ} {Ҹ} {ҹ} {Һ} {һ} {Ҽ} {ҽ} {Ҿ} {ҿ} {Ӏ} {Ӄ} {ӄ} {Ӆ}
  {ӆ} {Ӈ} {ӈ} {Ӌ} {ӌ} {Ӎ} {ӎ} {Ӕ} {ӕ} {Ә} {ә} {Ӡ} {ӡ} {Ө} {ө}
 }
 {
  \__recode:c { u8: \tl_to_str:n { #1 } }
 }
\ExplSyntaxOff

\title{Automatic Speech Recognition of Low-Resource Languages Based on Chukchi}

\author{
 Cydnie Davenport \\
  Linguistic Theory and Language Description Masters program\\
  Higher School of Ecomonics\\
  Moscow, Russia \\
  \texttt{davenport.cyd@gmail.com} \\
   \And
 Emil Nadimanov \\
  Computational Linguistics Masters program\\
  Higher School of Ecomonics\\
  Moscow, Russia \\
  \texttt{nadimaemi@gmail.com} \\
  \And
 Anastasia Safonova \\
  Computational Linguistics Masters program\\
  Higher School of Ecomonics\\
  Moscow, Russia \\
  \texttt{an.saphonova@gmail.com} \\
  \And
 Tatiana Yudina \\
  Computational Linguistics Masters program\\
  Higher School of Ecomonics\\
  Moscow, Russia \\
  \texttt{yudina.tatiana22@gmail.com} \\
}

\begin{document}
\maketitle


\section{Introduction}
The following paper presents a project focused on the research and creation of a new Automatic Speech Recognition (ASR) and Text to Speech (TTS) system based in the Chukchi language. The aim of this is to develop a system that makes the language more accessible to speakers of Chukchi - such as annotating subtitles on videos and movies, providing more accessible data for research and analysis, or the creation of chat-bots for online users. This system should consist of; an acoustic model for receiving an audio signal fragment and which gives the probability of various phonemes based on the fragment analyzed; a language model for determining which suggestions are more or less likely; and a decoder which will determine the most likely prediction. Predictive automatic speech recognition models already exist and are a popular focus in the realm of Natural Language Processing, however, the most challenging adversaries are low-resource languages due to extreme data deficits. 

This project is centered around a multi-step research process. Initially, we began by analyzing the Chukchi language from a linguistic perspective, but for the sake of clarity regarding motivations for making this system, it also must be looked at from a cultural and sociolinguistic perspective. What is known about the language, are there any cultural influences within the language, why is such a system necessary, and so on. Once there was extensive understanding of the subject at hand, the next step would be finding data that is usable. For this project, this included broadcasts from a Russian-based Chukchi radio station, videos and lessons from YouTube, written translations of the Bible, and the Higher School of Economics’ set of Chukchi-based corpora known as Chuklang. Once enough data is collected, there then comes the task of cleaning it. This included labeling and segmenting audio data for training, cleaning and filtering out unnecessary symbols (mainly Russian) from text, and determining which data would be used for pre-training and which would be used for testing the resultant model. Once enough data has been collected and cleaned for our model, we sample and train various models to understand how they process data. Additionally, we must try various encoders to understand how well they clean out noise and extra acoustic audio. Extra research must be conducted in order to compare models designed for both high- and low-resource languages. Various designs and tools for training ASR models include VQ-VAE, XLSR, the toolkit Kaldi, wav2vec, and more. The intended result of this project is an automatic speech-recognition system that can seamlessly work with Chukchi and provides us with the potential to be used for other low-resource languages.

\section{BACKGROUND}
\subsection{The Chukchi Language}
The Chukotko-Kamchatkan family of languages is said to contain two branches by default. The northern branch is referred to as the Chukotian branch (or “Luorevetlan”, based on the Chukchi ethnonym) and consists of Chukchi, Koryak, Alutor and Kerek (now extinct). The second branch is known as Itelmen, and contains the language Western Itelmen, which itself consists of two dialects: Khajrjusovo and Sedanka \cite{fortescue2005}. The language of focus for this paper is Chukchi, a polysynthetic language spoken primarily within the Chukotka Autonomous Okrug, which is located in the easternmost part of Siberia. Chukchi itself is an endangered indigenous language with less than 10,000 speakers at present, and most speakers are bilingual with a primary language of Russian. There are only less than 100 speakers who don’t speak Russian at all. Instances and usages of this language are difficult to come by, and is not a language taught in schools. The decreasing use of this language in general everyday life, as well prominence of Russian within the community demonstrates the necessity for an automatic speech recognition system, so that we may provide more accessibility to such an endangered and very low-resource language and its community. 
\subsection{What is a Low-Resource Language?}
In the field of NLP, research tends to have a large focus on languages where data and native speakers are easily accessible, and the language is relatively well-known. These are referred to as high-resource languages, and as such, produce a large quantity of data. On the other hand, low-resource languages (occasionally referred to as LRLs) are usually “..less studied, resource scarce, less computerized, less privileged, less commonly taught, or low density..” \cite{magueresse2020} and therefore are not prioritized in the realm of NLP research. However, this is actually one of the more major motivating factors for our project. Chukchi is an incredibly low-resource language, an example of which is that most of the up-to-date information regarding the language and its speakers is most easily accessed from a detailed article found on Wikipedia\footnote{Link to the article in question: \url{https://en.wikipedia.org/wiki/Chukchi\_language\#:~:text=Chukchi\%20\%2F\%CB\%88t\%CA\%83\%CA\%8Ak,mainly\%20in\%20Chukotka\%20Autonomous\%20Okrug.\&text=In\%20the\%20UNESCO\%20Red\%20Book,the\%20list\%20of\%20endangered\%20languages}}. The low-resourcedness of Chukchi is what inspired this project, as it is an endangered language, and one that is not particularly accessible in terms of media, education, and history. By creating a new automatic speech recognition system, not only can accessibility be provided for this language, but it also creates new opportunities for the same achievements in other low-resource languages.

\subsubsection{What is an ASR System?}
Traditionally, modern automatic speech recognition systems are typically made up of three different parts: a lexicon, an acoustic model, and a language model\footnote{Information about the basic components of an automatic speech recognition system is widely available, one of the more easily understandable sources can be found here: \url{https://voximplant.com/blog/what-is-automatic-speech-recognition}}. The lexicon contains the information that an ASR system needs to be able to understand the input it receives on the base level. This includes things such as phonetic transcription codes that are used for the target language’s phonemes. For English, ARPABET\footnote{\url{https://en.wikipedia.org/wiki/ARPABET}} and TIMIT\footnote{\url{https://en.wikipedia.org/wiki/TIMIT}} are the most commonly used codes and transcriptions, developed by the Defense Advanced Research Projects Agency (DARPA). 

The second component of an ASR system is the acoustic model, which is responsible for forging the relationships between the phonemes of a language (such as the ones provided in the lexicon) and an audio signal. This interaction is supported by the use of transcripts along with their respective audio files\footnote{Microsoft conducts extensive research regarding acoustic models, more information as well as links to other sources and publications can be found here: \url{https://www.microsoft.com/en-us/research/project/acoustic-modeling/}}, and are thus supposed to be able to map statistical representations for feature vector sequences of a particular phoneme (or sound unit) and classify it \cite{sarma2015}. This allows the system to recognize and distinguish this particular sound unit from the rest of the phonemes that it may encounter in both training data and experimental data. 

Finally, there is the language model, which helps to provide clearer contexts and allows the model to view the language in a naturally occurring form. Thus, this is where training comes in. By training the language model, contexts become more comprehensive and coherent when interacting with the system, and are thus understandable. By design, the system, with the help of all of these aforementioned components, is then supposed to be able to predict speech patterns.

\subsection{Previous Research}
Regarding previous research focused on low-resource languages and ASR, there have been multiple approaches to finding the most efficient and effective model for processing such a limited amount of available data. The basic framework for processing speech was typically based on a few components. For example, these components could have included an autoencoder (denoising or otherwise), dual transformation for both text and speech, bidirectional sequence modeling, typically with a major focus on unsupervised pre-training \cite{ren2020}. In addition to this, many approaches also included a Transformer-based unified model structure \cite{ren2020}, \cite{kriman2020}. The goal was to have a system that could sample the language evenly and return feedback to the model, learning as it continued to sample more data \cite{yubei2021}. These components are crucial to the creation of our model, and will be utilized in this project.

Both universities and major corporations alike (e.g., Google with Strope et al, 2011) have also researched the most effective ways to implement the most ideal features for training both acoustic models as well as language models. In many cases, it is incredibly difficult to create a pre-training environment that is entirely unsupervised, but the key here is that it is almost unsupervised. The benefit of unsupervised data pre-training is that it makes data much more usable. Without the need for supervision, the amount of usable data increases significantly, which gives us much more accessibility to languages that lack a significant amount of data (i.e., being able to utilize data in a more efficient way). The key to much of the unsupervised training that already exists is the technique implemented. Discriminative, in which a dual unigram and trigram language model was used to interpret relative truth, active or passive. Passive learning was a technique and algorithm that dominated the realm of automatic speech recognition for much of its lifespan.

Passive learning was the initial algorithm used for training language models. This meant that a model was trained based on a single implementation of a set of data, fixed in time. As a result, there was no room allowed for the model to improve. Additionally, all the data being used was usually transcribed under human supervision, and training a model to work with language data ended up being a very time consuming process. By taking the workload off of the researchers and volunteers who transcribe this data and manually check the model, more effective and efficient means of training an ASR system can be developed. Active learning mechanisms, as a result, are particularly useful in cases like these. With a feedback system that allows for the model to learn from itself and ultimately use less data. This can prove invaluable in developing an ASR system for Chukchi and other low-resource languages.


\section{DATA COLLECTION}

Given that Chukchi is a very low-resource language with very few speakers, finding usable data proved difficult, as was discussed above. Samples of both spoken and written Chukchi were selected from any source that could be found. This included the Charles Weinstein website of Chukchi with translations and descriptions in both French and Russian, recent news broadcasts in Chukchi (December 2020 and January 2021) from the Anadyr’-based radio station Radio “Purga”, videos from Youtube, corpora from Chuklang.ru, as well as translated parts of the Bible from Bible.is.

\subsection{Radio ‘Purga’}
One of the main sources of high-quality annotated data was the Chukchi radio station Radio “Purga,” which has a special feature at their station in which they report news in Chukchi on a regular (almost daily) basis. A representative of this radio station provided our research team with a total of 2.53 hours of audio data from 30 episodes of morning news. These audio files then had to be manually split into shorter chunks of both audio recording and text pairs in order to be used further. Fortunately, each broadcast came with its own script. However, there were a few issues with this data in that the real recording and the script would sometimes differ. Additionally, the script also contained several sentences of pure Russian speech, which makes some part of the audio file unusable. Both of these issues were solved by excluding such recording-script pairs from the dataset.

\subsection{YouTube Videos}
Youtube was another primary resource to find any instances of Chukchi audio samples. All videos found were then converted into WAV format. This portion of the corpus contains:
\begin{itemize}
\item stories; 
\item Chukchi online language lessons; 
\item interviews with native speakers; 
\item lessons from the project “Vetgav. Chuckhi lessons”; 
\item cartoons; 
\item news; 
\item and more. 
\end{itemize}
All links to the video data used for this project and statistical sources can be found at the project’s GitHub\footnote{Project GitHub: \url{https://github.com/ftyers/fieldasr/blob/main/DATA.md}}. The video data of the corpus totaled at 14 hours, 12 minutes and 13 seconds.

\subsection{Bible Audio}
The resource Bible.is\footnote{Bible.is: \url{https://www.faithcomesbyhearing.com/audio-bible-resources/bible-is}} contains chapters from the Bible in a variety of languages from around the world, including Chukchi. There we found audio recordings totaled at 3 hours, 36 minutes and 52 seconds in wav format. Some Bible chapters do have a text annotation (for example, The Gospel of Luke), and some don’t (The Book of Jonah).

\subsection{Chuklang Corpora}
The Chuklang Corpora\footnote{Chuklang corpora: \url{https://chuklang.ru}} was created by professors and students of the National Research University - Higher School of Economics in Moscow, Russia, during the linguistic expeditions to the village of Amguema in the Iultinsky District of the Chukotka Autonomous Okrug. It consists of annotated audio recordings with a total length of 1 hour, 14 minutes and 18 seconds. 

\subsection{All audio data}
Table~\ref{tab:table1} displays information about the duration of all audio data, broken down by resource type. \\

\begin{table}[h]
 \caption{Distribution of audio}
  \centering
\begin{tabular}{| p{2cm} | p{2cm} | p{2cm} |}
    \hline
    Resource          & Duration & Transcription availability \\
    \hline
    Radio ‘Purga’     & 2:32:00  & +                          \\
    \hline
    Youtube           & 14:11:13 & -                          \\
    \hline
    Bible             & 3:36:52  & -                          \\
    \hline
    Chuklang          & 1:14:18  & +                          \\
    \hline
\end{tabular}
\label{tab:table1}
\end{table}
Duration of all unannotated data - 17:48:05 \\
Duration of all annotated data - 3:46:18 \\
Total duration of all audio data - 21:34:23 \\

\subsection{Text}
For a linguistic model in an automatic speech recognition system, any kind of texts are useful, even if they do not have their own audio annotations. The most commonly used lexical items (ынкъам, гивик, ымы, ытлён, ынан, Чукоткакэн, гатвален, ынӄэн, лыги, вальыт, вагыргыттитэ, гатваленат) were used to assist in finding written examples of Chukchi. In addition to this, Yandex.XML\footnote{Yandex.XML: \url{https://yandex.ru/dev/xml/}}, which is a service that allows a user to send a search query to the Yandex search engine and receive any answers found in XML format, was also used. No more than 200 results per query were allowed per its own limitations, however, by using various filters and sorting mechanisms, 1800 URLs were found and 462 unique links were extracted. From these selected URLs, the largest and most useful sources for parsing were determined to be:
\begin{itemize}
\item news outlets
\item stories and riddles
\item fictional literature
\item grammar and thematic dictionary of the Chukchi language, a collection of Chukchi literary texts
\end{itemize}
	
 The full text corpus is composed of 112,719 sentences, and 2,068,273 words. The obtained URL links and text corpus can be found on the GitHub repository for this project\footnote{Project GitHub: \url{https://github.com/ftyers/fieldasr/blob/main/DATA.md}}. Table~\ref{tab:table2} below displays the distribution of texts according to subcorpus. From these numbers, the necessity for automatic deletion of Russian from a few sub corpora becomes quite noticeable.

\begin{table}[h]
 \caption{Distribution of texts by subcorpus (before the removal of the Russian language)}
  \centering
\begin{tabular}{| p{5cm} | p{2cm} | p{2cm} | p{2cm} | p{1.5cm} | p{1.5cm} |}
    \hline
    Subcorpus & Number of texts/ pages & Number of sentences & Number of words & \% of words & Russian language presence \\
    \hline
    Internet newspaper "Extreme North" & 118 & 6187 & 82661 & 4,00\% & + \\
    \hline
    A special supplement to the newspaper "Extreme North" & 11 & 303 & 27466 & 1,33\% & + \\
    \hline
    "Portal of National Literatures" & 4 & 569 & 4729 & 0,23\% & - \\
    \hline
    The book "По аргишному пути канчаланского чаучу" & 1 & 444 & 3936 & 0,19\% & + \\
    \hline
    Lingvoforum: fairy tales, stories & 8 & 251 & 1297 & 0,06\% & - \\
    \hline
    VK: fairy tales & 11 & 573 & 2840 & 0,14\% & - \\
    \hline
    Puzzles & 1 & 1234 & 8112 & 0,39\% & + \\
    \hline
    The Wayback Machine & 47 & 3079 & 16952 & 0,82\% & - \\
    \hline
    Charles Weinstein & 10 & 96765 & 1891705 & 91,46\% & + \\
    \hline
    Radio ‘Purga’ & 29 & 836 & 7742 & 0,37\% & + \\
    \hline
    Chuklang Corpora & 1006 & 1006 & 4414 & 0,21\% & - \\
    \hline
    Bible.is & 25 & 1472 & 16419 & 0,79\% & - \\
    \hline
    All data & 1271 & 112719 & 2068273 & 100\% & \\
    \hline
\end{tabular}
\label{tab:table2}
\end{table}

\section{DATA PROCESSING}
\subsection{Preprocessing: Text}
The Chukchi alphabet contains a special symbol (‘) which was implemented in different forms in writing (‘, ’, `, “, ”, etc) across different sources of Chukchi. Therefore, the first step for preprocessing was to identify all instances of this apostrophe and determine a single variant for its representation, with the result being: (’). Following this were substitutions of letters к’ - ӄ, K’- Ӄ, н’ - c, Н’ - Ӈ, where the rightmost form is the final representation needed.

Additionally, we cleaned up the data collected from internet-sources on a more global scale: that is, deletion of hyperlinks, illegible symbols, email addresses (these can frequently be found in news articles), and occasionally, double spaces between words. All data preprocessing was carried out with the help of the standard Python library re\footnote{Documentation on the Python library re can be found here: \url{https://docs.python.org/3/library/re.html}}, which is used for work with regular expressions.

\subsection{Removing Russian from the Text Corpus}
Due to the writing system of both Chukchi and Russian being incredibly similar, it became necessary to clean the Russian out of the text corpus. This is because the similarities between the visible forms of the languages can be problematic to the model, and thus skew our results. Binary sentence classification with the help of the most frequently used words in both Russian and Chukchi. Based on the collected text data that only contains Chukchi, we calculated the most frequently occurring words. The “New Russian Lexical Frequency Dictionary”\footnote{New Russian Lexical Frequency Dictionary: \url{http://dict.ruslang.ru/freq.php}} was used to create a list of the most frequently occurring Russian words. In order to predict which class a sentence belonged to, each word in the text, along with its normal form determined by the morphological analyzer pymorphy2\footnote{Documentation for pymorphy2: \url{https://pymorphy2.readthedocs.io/en/stable/}}, were checked for entries in the frequency list. The corresponding variable-counters were increased and then compared. If a Russian word was found more than a Chukchi word, then that sentence was considered Russian, and vice versa.

The pre-training model “lid.176.bin” was used for fastText\footnote{Documentation for fastText: \url{https://fasttext.cc/}} for language recognition. This model is supported for 176 languages. With the assistance of this model, the likelihoods of sentences belonging to certain languages were isolated, followed by the use of the model KMeans\footnote{Documentation for KMeans: \url{https://pymorphy2.readthedocs.io/en/stable/}} for segmenting data into 2 clusters: Chukchi and not-Chukchi. For the given model the f1-score value was 0.91. After deletion of Russian from the corpus, the size of the resulting corpus was 15309 sentences and 117567 words. This corpus can be found on the GitHub repository for this project\footnote{Project GitHub: \url{https://github.com/ftyers/fieldasr/blob/main/data/text_corpus.txt}}.

\begin{table}[h]
 \caption{Distribution of texts by subcorpus (after the removal of the Russian language}
  \centering
\begin{tabular}{| p{5cm} | p{2cm} | p{2cm} | p{2cm} | p{1.5cm} | p{1.5cm} |}
    \hline
    Subcorpus & Number of texts/ pages & Number of sentences & Number of words & \% of words \\
    \hline
    Internet newspaper "Extreme North" & 118 & 3063 & 33219 & 28,26\% \\
    \hline
    A special supplement to the newspaper "Extreme North" & 11 & 3236 & 25160 & 21,40\% \\
    \hline
    "Portal of National Literatures" & 4 & 574 & 4732 & 4,02\% \\
    \hline
    The book "По аргишному пути канчаланского чаучу" & 1 & 208 & 1425 & 1,21\% \\
    \hline
    Lingvoforum: fairy tales, stories & 8 & 304 & 1297 & 1,10\% \\
    \hline
    VK: fairy tales & 11 & 755 & 2840 & 2,42\% \\
    \hline
    Puzzles & 1 & 358 & 1511 & 1,29\% \\
    \hline
    The Wayback Machine & 47 & 3127 & 16904 & 14,38\% \\
    \hline
    Charles Weinstein & 10 & 332 & 1989 & 1,69\% \\
    \hline
    Radio ‘Purga’ & 29 & 863 & 7654 & 6,51\% \\
    \hline
    Chuklang Corpora & 1006 & 1006 & 4416 & 3,76\% \\
    \hline
    Bible.is & 25 & 1483 & 16420 & 13,97\% \\
    \hline
    All data & 1271 & 15309 & 117567 & 100\% \\
    \hline
\end{tabular}
\label{tab:table3}
\end{table}

\subsection{Attempts at Denoising}
Due to Chukchi being a very low-resource language, the amount of high quality audio recordings in our dataset were minimal. Most of the audio that dominated the collected data were those of fairly poor quality. In order to achieve better results with the data on hand, a denoising autoencoder was applied to the data to filter out extra noise.

However, either due to flaws in denoising methodology or due to the high level of noise pollution in the collected audio data, the application of a denoising autoencoder was fruitless, and did not appear to aid our system. In the future, we would like to find something that will clean our audio in a more efficient and effective way.

\subsection{Segmenting and Labeling Radio Data}
The samples from Radio “Purga” were collected for the months of December 2020 and January 2021. These files typically consisted of a short news segment, usually no longer than 5 minutes each. With about 15 of these samples collected for each month, we had 200 minutes of audio. However, these audio files needed to be segmented, as it was unfortunately not all Chukchi. These audio recordings were segmented by sentences, and labeled using the software ELAN\footnote{ELAN is a linguistic annotator that allows for work with audio files, including segmentation, and labeling.}. Any instances of Russian were to be extracted and removed from the files. In addition to this, the audio broadcasts were also provided with personal scripts from the speaker’s notes. As a result, we were able to take these scripts, and segment the audio by sentences, which was both necessary for processing data and training the model. However, some of these scripts appeared to be used more as notes as opposed to actual scripts, with some portions of the broadcast entirely missing from the documents we had been provided with. In these cases, the audio was not usable, and was removed from the dataset.

\subsection{Audio Data Slicing}
All received audio was cut into small fragments. Sometimes it was possible to cut by pauses, so as not to cut off parts of words that may be important for ASR model training. In this case, the Python libraries pyAudioAnalysis\footnote{\url{https://github.com/tyiannak/pyAudioAnalysis}} and pydub\footnote{\url{https://github.com/jiaaro/pydub}} were used. Sometimes it was impossible to cut by pauses (for example, if there was music playing in the background), if this was the case, then the files were cut by 5 minutes.


\section{METHODOLOGY}
This project was centered around 3 experiments with 3 different models: VQ-VAE, XLSR, and wav2vec. These models were selected specifically for their previously displayed use for low-resource languages. Each model was pre-trained with a portion of the cleaned, segmented, annotated and unannotated Chukchi data, and then tested with the remaining portion. 

\subsection{VQ-VAE}
\subsubsection{Theory}
One approach used in creation of ASR systems for low-resource languages is to use a model that can learn from latent representations in language. For this task the best option would be to use VQ-VAE  \cite{oord2018}. This model has two important components: VAE (Variational Auto-Encoders) \cite{kingma2014} and VQ (Vector Quantized). VAE is a type of NN architecture which does inference by retrieving statistics (mean and variance) for latent random variables. Model learns separate iand ifor a random variable. VAE works for continuous data, а вот VQ-VAE learns discrete representations (embeddings). It is assumed that the model VQ-VAE will learn embeddings by constructing its own inputs. Therefore, only audio data is required for learning. Thus, if the learned latent representations are learned well, then they can be used for unsupervised or semi-supervised models.

\begin{figure}[h]
  \centering
  \includegraphics[width=1\linewidth]{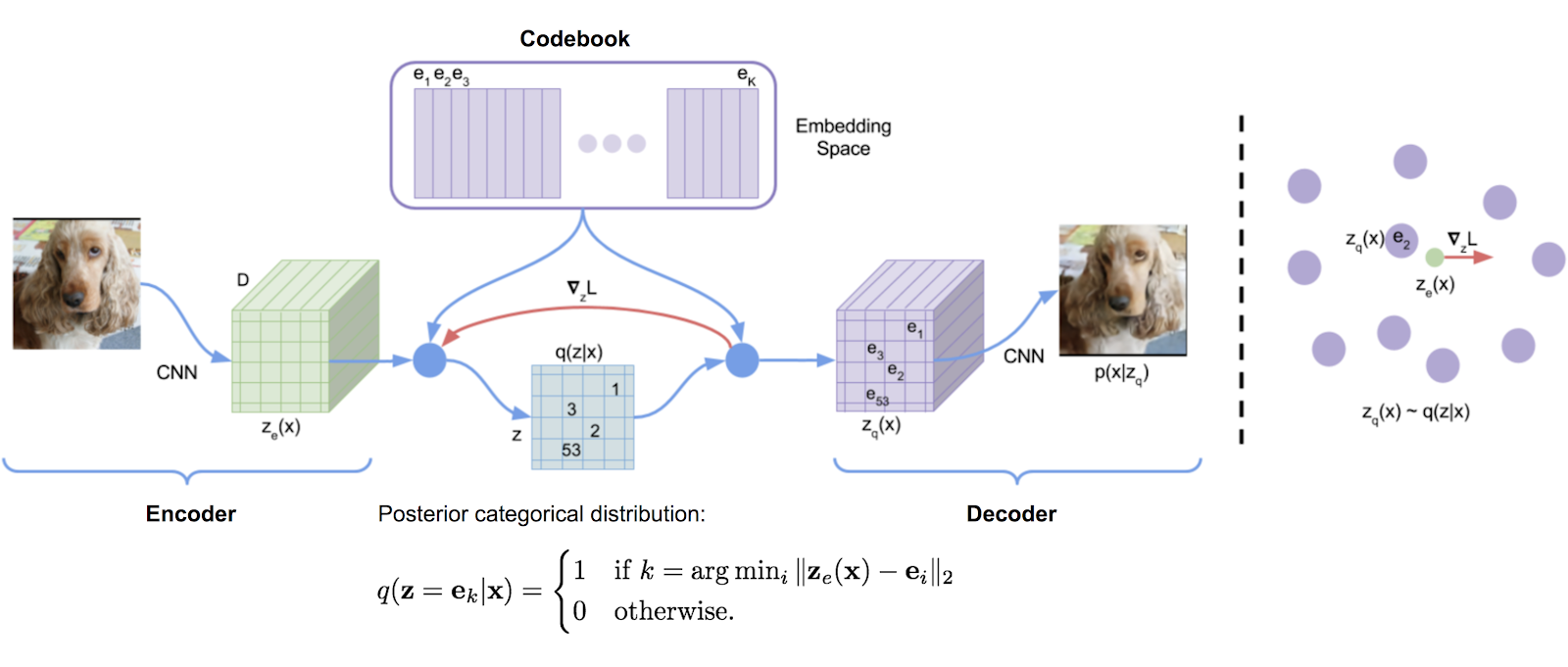}
  \caption{Left: A figure describing the VQ-VAE. Right: Visualisation of the embedding space. (Oord, Vinyals, et al. 2017)}
  \label{fig:fig1}
\end{figure}

\subsubsection{Experiment}
Two experiments were carried out for the collected audio data: Chuklang, Bible.is, Radio “Purga,” and YouTube. The data corpus used in the first experiment contained audio with a length of 4-5 minutes (as a result a large part of this audio was cut down to 5-20 seconds), therefore a decision was made to conduct a second experiment in which the audio that had lengths of longer than 20 seconds were converted into smaller audio fragments, segmented by pauses. This was done as the length of the largest piece of audio in the dataset influences the padding parameter value of the model: the model can have a significant bias in the data, as it will fill all the audio-recording vectors with zeros to fit the size of the largest audio piece.

The second experiment that was carried out focused on the use of data augmentation. We attempted to apply various effects from the first experiment to the audiodata, thereby doubling the size of the training dataset. Regarding effects, the following were used:
\begin{itemize}
\item single-pole lowpass filter (["lowpass", "-1", "300"]),
\item reduce the speed (["speed", "0.8"]),
\item after that it was necessary to add the `rate` effect with the original sample rate (["rate", f"\{sample\_rate1\}"]),
\item reverberation (["reverb"\, "\-w"]). 
\end{itemize}

In the present work we used the model VQ-VAE for PyTorch. The following parameters were used for the VQ-VAE model:
\begin{itemize}
\item input\_dim=39,
\item hid\_dim=256,
\item enc\_dim=64,
\item K=512. 
\end{itemize}

The VQ-VAE model was trained for 1000 epochs. The following values were selected as the batch-size for the data: 128 (train), 10 (validation), 16 (test). PyTorch’s Adam Optimizer was also implemented with a learning rate of 2e-4.

\subsubsection{Results}
The validation loss function was used in order to rate the quality of the VQ-VAE model. The results from the change in the function can be observed on Table~\ref{tab:table4}. \\

\begin{table}[h]
 \caption{Validation loss values for experiments with VQ-VAE}
  \centering
\begin{tabular}{| p{3cm} | p{3cm} | p{3cm} |}
    \hline
    Experiment & Validation loss (1 epoch) & Validation loss (1000 epoch) \\
    \hline
    Experiment 1.1 (without augmentation) & 3249.4058 & 3132.565 \\
    \hline
    Experiment 1.2 (with augmentation) & 3247.8926 & 3131.7913 \\
    \hline
    Experiment 2 & 3249.5176 & 3128.0374 \\
    \hline
\end{tabular}
\label{tab:table4}
\end{table}

It is visible from this Table~\ref{tab:table4} that even after 1000 epochs, the value of the validation loss was left practically unchanged. As a result, we have determined that this model was unsuccessful for our research purposes due to 2 main reasons: the small amount of data available for training, and the poor sound-quality of these audio files. As mentioned previously, the denoising autoencoder proved ineffective, resulting in audio that still contained a significant amount of noise, likely interfering with the model.

\subsection{XLSR}
\subsubsection{Theory}
Another approach to this task is using a model trained on a high-resource language and fine-tuning it via low-resource language data, in our case, Chukchi. A model that is suitable for such a cross-language approach is referred to as XLSR or Cross-Lingual Speech Representations. XLSR is a transformer-based multilingual model that was trained on 56 thousand hours of speech data shared between 53 languages. It is supposed to represent latent features that are shared between languages. This means that XLSR can be used as a solid base for fine-tuning for low-resource languages. Therefore, Chukchi, being extremely low-resource, is a great candidate for this experiment. For a broader description of the model and this approach we suggest addressing the original article by Facebook AI team \cite{conneau2020}. 

We hypothesized that through using a model that has learned generalized natural language representation, it will be easier to train a model that recognizes Chukchi speech - basically, a big part of the training process is skipped, and the whole training process is essentially condensed to fine-tuning. Moreover, the authors of the mentioned article have proven that multilingual models may even outperform monolingual ones for some languages.

\begin{figure}[h]
  \centering
  \includegraphics[width=1\linewidth]{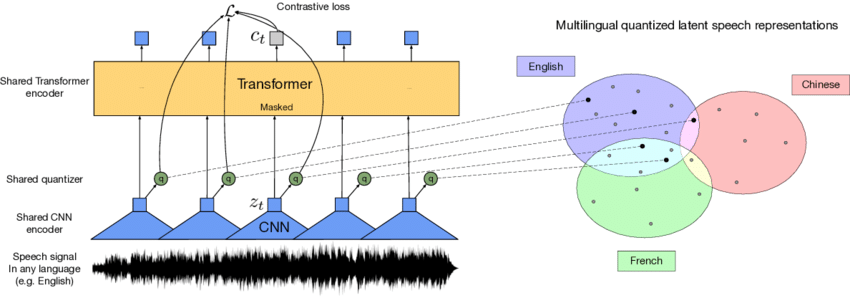}
  \caption{A shared quantization module over feature encoder representations produces multilingual embeddings, through which the model learns to share discrete tokens across languages, creating bridges across languages \cite{conneau2020}}
  \label{fig:fig2}
\end{figure}

\subsubsection{Experiment}
After the necessary initial steps, such as feature extraction, normalization, sampling rate adjustment, etc., the model was further trained on the annotated Chukchi data at hand. The CNN part from Figure 2 in section 5.2.1 does not need fine-tuning, because it was trained by the authors of the model where the multilingual features are encoded. Moreover, the fine-tuning process is relatively short, and only lasts for 6 epochs, 400 steps each, because the model has shown to be prone to overfitting on such a small dataset. The training rate was set to a static value of 3e-4, which fits in the range suggested by the authors of the original paper. It seems to make little sense to change learning rate dynamically, given the low amount of training data.

The model will be publicly available at our team’s GitHub repository\footnote{Project GitHub: \url{https://github.com/ftyers/fieldasr/blob/main/DATA.md}}.

\subsubsection{Results}
The data that was used for fine-tuning consisted of annotated data that had the structure shown on Table~\ref{tab:table5}.

\begin{table}[h]
 \caption{Training and Testing times for the XLSR model}
  \centering
\begin{tabular}{| p{3cm} | p{3cm} |}
    \hline
    TRAIN & 02:49:36.35 \\
    \hline
    TEST & 00:17:09.25 \\
    \hline
    TOTAL & 03:06:45.00 \\
    \hline
\end{tabular}
\label{tab:table5}
\end{table}

This allowed the fine-tuned model to achieve a WER\footnote{WER is a metric that is commonly used for speech modelling and speech recognition. It is computed at word-level as (Deletions + Insertions + Substitutions) / N, where N is the number of words in a text, and the three components of the dividend are the numbers of corresponding operations, applied to words, that are needed to reconstruct the original text.} of 0.758395 and CER\footnote{CER is computed the same way as WER, but on character level. That means that the same formula is used and the same operations are counted, but in appliance to characters in the original text.} 0.186895 - surprisingly low for such a small dataset. We plan to prepare more data and further improve this result, as it has appeared to be the most fruitful approach among others. Additionally, it appears to be useful to compare these metrics with the results achieved in the original article (see Table~\ref{tab:table6}).

\begin{table}[h]
 \caption{Word Error Rate}
  \centering
\begin{tabular}{| p{3cm} | p{1.5cm} | p{1.5cm} | p{1.5cm} | p{1.5cm} | p{1.5cm} |}
    \hline
    Language & Assamese & Tagalog & Swahili & Georgian & Our: Chukchi \\
    \hline
    WER & 44.1 & 33.2 & 26.5 & 31.1 & 75.8 \\
    \hline
    Fine-tuning data, hrs & 55 & 76 & 30 & 46 & 3 \\
    \hline
\end{tabular}
\label{tab:table6}
\end{table}

WER proved to not be the best metric for training in Chukchi, as it accounts for any mistake in a word. Considering the nature of this language is polysynthetic (e.g. it contains long “sentence-words”), it is easy to achieve a high error rate due to a large number of substitutions. CER is, in our opinion, a more suitable metric in these circumstances. Expectedly, in comparison with the results in the Table~\ref{tab:table7} above, our model does not perform satisfactorily on the word-level.

\begin{table}[h]
 \caption{Character Error Rate}
  \centering
\begin{tabular}{| p{3cm} | p{1.5cm} | p{1.5cm} | p{1.5cm} | p{1.5cm} | p{1.5cm} |}
    \hline
    Language & Assamese & Tagalog & Swahili & Lao & Ours: Chukchi \\
    \hline
    CER & 17.9 & 13.1 & 21.3 & 22.4 & 18.7 \\
    \hline
    Fine-tuning data, hrs & 55 & 76 & 30 & 59 & 3 \\
    \hline
\end{tabular}
\label{tab:table7}
\end{table}

Character error rate accounts for character-level mistakes. Our model has achieved a comparable result, given much less data. This result is very satisfying, however, low WER definitely means that this model is not suitable for production use. However, if one takes into consideration the morphological complexity of Chukchi and its incredibly scarce data, one can conclude that XLSR has proven to be an astoundingly powerful base for low-resource ASR.

As a demonstration, we would like to finish this part of the article with an example of the model’s output on Table~\ref{tab:table8}

\begin{table}[h]
 \caption{Output from the XLSR model}
  \centering
\begin{tabular}{| p{5cm} | p{5cm} |}
\hline
Original & Recognised \\
\hline
ӈутингивик ымыӆьычукоткак ӈыраӄ аӄатвагыргын гатваленат яама нымытваӆьа милгэрти ыннэнчьэн о'равэтӆьан егтэлытваркын ӄутти ӈыроӄ гэвъилинэт & ӈутингивик м чукуткак ӈыаӄалтвагыргыт гатваленат яма нымытваӆьа милгэри ынэчьэнноравэтлан егтэлвркын отиӈыргэвилиэт \\
\hline
\end{tabular}
\label{tab:table8}
\end{table}

\subsection{Wav2Vec Unsupervised}
\subsubsection{Theory}
\begin{figure}[h]
  \centering
  \includegraphics[width=1\linewidth]{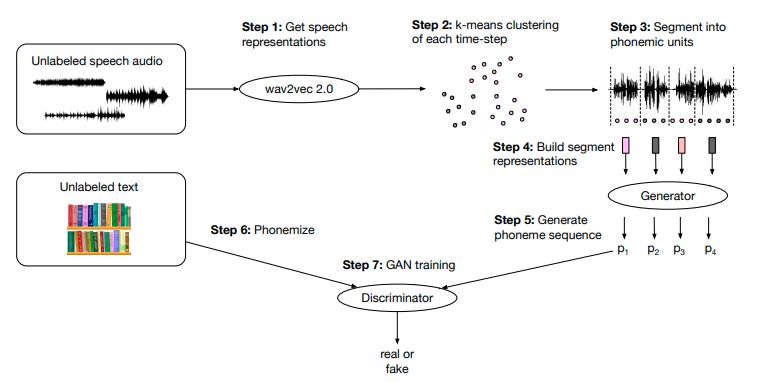}
  \caption{Illustration of wav2vec Unsupervised (Baevski, et al. 2021)}
  \label{fig:fig3}
\end{figure}

Another model suitable for our task, wav2vec Unsupervised (wav2vec-U), was presented in May 2021 by Facebook AI \cite{baevski2021}. The main feature of this model is that only separate audio and texts that are not related to each other are needed (not annotated ones), which makes it possible to apply it to low-resource languages that lack text and audio pairs. A good result (about 85\% for low-resource languages, such as Tatar) is obtained due to  the self-supervised model (wav2vec 2.0), a simple k-means clustering method and a Generative Adversarial Network (GAN). The illustration of wav2vec-U can be found in Figure~\ref{fig:fig3}. The wav2vec-U training procedure consists of three main steps:\\
\begin{itemize}
\item Preparation of speech representations and text data;
\item Generative Adversarial Training (GAN);
\item Iterative self-training + Kaldi LM-decoding.
\end{itemize}

\subsubsection{Experiment}
As mentioned above, using the model requires specific preprocessing of the data, which requires wav2vec 2.0 (open-source framework for self-supervised learning of representations from raw audio data), eSpeak (a compact open source software speech synthesizer for English and other languages) and fastText (a library for efficient learning of word representations and sentence classification) pre-trained models for Chukchi, which do not exist. It is possible to train these models from scratch, however, it takes a lot of time, and at the moment of writing this article it was not possible to preprocess the data properly for the wav2vec-U model.

\subsubsection{Results}
No results for wav2vec-U have been obtained at this time, this will be a task for the next study.


\section{CONCLUSION}
In this work we introduced our project, focused on creating a new automatic speech recognition (ASR) system for low-resource languages, with a focus on Chukchi. Our goal was to create a system as unsupervised as possible. In order to run experiments and train our selected models to work for Chukchi, we collected a sizable corpus of both audio and text data in Chukchi, a feat that was rather unexpected for such a low-resource language. In total, there were 15,309 sentences, and 117,567 words for the full text corpus, and 21 hours, 34 minutes, 23 seconds worth of audio data. We then proceeded to conduct 3 different experiments centered around VQ-VAE, XLSR, and wav2vec models. 

The first experiment, regarding Vector Quantized - Variational Auto-Encoder (VQ-VAE), demonstrated very little change in validation loss between the first and thousandth epoch that had been executed. This result was found in all three sub-experiments that were run with both augmented and unaugmented data. This minimal change is likely a result of how little data we had for this model, as well as the poor sound quality of the audio files. The second experiment, focused on XLSR, was shown to be more promising than the first, with a surprisingly low WER of 0.758395 and CER 0.186895. XLSR actually proved to be the most powerful base for development in low-resource ASR. These results on their own may not appear particularly noteworthy, however, we must take into account the fact that Chukchi is a polysynthetic language, meaning that words and sentences tend to have more of an overlapping appearance when manifested in written language. With this in mind, our interpretation of the CER result becomes much more significant. Finally, the third experiment conducted through wav2vec, unfortunately, proved fruitless. There are no results to interpret at this time, as preprocessing of the data was unsuccessful. This experiment will be pursued in further studies.
 
In summary, we were able to train 2 separate models for work in Chukchi, and will be attempting to train the third in the near future. Despite some unfortunate setbacks, we have made an incredible amount of progress in our understanding of low-resource automatic speech recognition, have learned what works for this particular set of data and what proves ineffective, and set tasks that can be completed in future research. Ultimately, we consider this to be a successful project with a promising outlook on the future of low-resource automatic speech recognition.

\bibliographystyle{unsrt}  


\end{document}